\pdfoutput=1
\documentclass[letterpaper, 10 pt, journal, twoside]{IEEEtran}

\usepackage{graphicx}
\usepackage{multirow}
\usepackage{textcomp}
\usepackage{amssymb,amsmath}
\usepackage{booktabs}
\usepackage[caption=false,font=normalsize,labelfont=sf,textfont=sf]{subfig}
\usepackage[linesnumbered,ruled,vlined]{algorithm2e}
\usepackage{siunitx}
\usepackage{physics}
\usepackage[table]{xcolor}

\let\oldnl\nl
\newcommand{\nonl}{\renewcommand{\nl}{\let\nl\oldnl}}

\begin{document}

\title{QT-TDM: Planning With Transformer Dynamics Model and Autoregressive Q-Learning}

\author{Mostafa Kotb$^{1,2,*}$, Cornelius Weber$^{1}$, Muhammad Burhan Hafez$^{3}$, and Stefan Wermter$^{1}$%
\thanks{Manuscript received: July, 21, 2024; Revised October, 8, 2024; Accepted November, 4, 2024.}
\thanks{This paper was recommended for publication by Editor Jaydev P. Desai upon evaluation of the Associate Editor and Reviewers' comments.}%
\thanks{This work was supported by the German Research Foundation DFG under project CML (TRR 169) and Mostafa Kotb is funded by a scholarship from the Ministry of Higher Education of the
Arab Republic of Egypt.} 
\thanks{$^{1}$Mostafa Kotb, Cornelius Weber and Stefan Wermter are with the Knowledge Technology Group, Department of Informatics, Universität Hamburg, 22527 Hamburg, Germany
        {\tt\footnotesize \{mostafa.kotb, cornelius.weber, stefan.wermter\}@uni-hamburg.de}}%
\thanks{$^{*}$Corresponding author: Mostafa Kotb}
\thanks{$^{2} $Mostafa Kotb is also with Mathematics Department, Faculty of Science, Aswan University, 81528 Aswan, Egypt
        {\tt\footnotesize m.kotb@sci.aswu.edu.eg}}%
\thanks{$^{3} $Muhammad Burhan Hafez is with School of Electronics and Computer Science, University of Southampton, Southampton SO17 1BJ, UK
        {\tt\footnotesize burhan.hafez@soton.ac.uk}}%
\thanks{The code is available at https://github.com/2M-kotb/QT-TDM/tree/main}%
\thanks{Digital Object Identifier (DOI): see top of this page.}
}

\markboth{IEEE Robotics and Automation Letters. Preprint Version. Accepted November, 2024}
{Kotb \MakeLowercase{\textit{et al.}}: QT-TDM: Planning With Transformer Dynamics Model and Autoregressive Q-Learning}

\maketitle

\begin{abstract}
Inspired by the success of the Transformer architecture in natural language processing and computer vision, we investigate the use of Transformers in Reinforcement Learning (RL), specifically in modeling the environment's dynamics using Transformer Dynamics Models (TDMs). We evaluate the capabilities of TDMs for continuous control in real-time planning scenarios with Model Predictive Control (MPC). While Transformers excel in long-horizon prediction, their tokenization mechanism and autoregressive nature lead to costly planning over long horizons, especially as the environment's dimensionality increases. To alleviate this issue, we use a TDM for short-term planning, and learn an autoregressive discrete Q-function using a separate Q-Transformer (QT) model to estimate a long-term return beyond the short-horizon planning. Our proposed method, QT-TDM, integrates the robust predictive capabilities of Transformers as dynamics models with the efficacy of a model-free Q-Transformer to mitigate the computational burden associated with real-time planning. Experiments in diverse state-based continuous control tasks show that QT-TDM is superior in performance and sample efficiency compared to existing Transformer-based RL models while achieving fast and computationally efficient inference. 
\end{abstract}


\IEEEpeerreviewmaketitle

\section{Introduction}

\IEEEPARstart{L}{earning} an accurate predictive model of environment dynamics \cite{mbrl} is a challenging yet promising technique in Deep RL to enhance sample efficiency \cite{sample-efficiency1}, \cite{sample-efficiency2}, \cite{sample-efficiency3} and achieve generalization \cite{generalization1}, \cite{generalization2}, \cite{generalization3}. The Transformer architecture \cite{gpt} is a strong candidate for dynamics modeling, as it proves to be an excellent sequence modeler and shows outstanding performance across various domains, including Natural Language Processing \cite{transformer-xl}, Computer Vision \cite{vit}, and Reinforcement Learning \cite{chen2021decision}.

Transformer dynamics models (TDMs) \cite{generalistTDM}, \cite{iris} have proven effective in \textit{background planning} \cite{dreamer} scenarios, where an actor-critic model is trained on the imagined trajectories generated by the learned dynamics model. During inference, the learned actor-critic model selects the suitable actions. TDMs show an outstanding performance in discrete action spaces \cite{iris}, \cite{twm} and in long-term memory tasks \cite{transdreamer}.

In \textit{real-time planning} scenarios, where the learned dynamics model plans ahead by being unrolled forward from the current state to select the best action, TDMs encounter hurdles. Specifically,  inference is slow and computationally inefficient  \cite{generalistTDM}, \cite{janner2021offline} due to the autoregressive token prediction and the per-dimension tokenization scheme, which increases sequence length as the environment's dimensionality increases. This makes planning for long horizons impractical, especially in the robotics domain, where fast inference is essential. Therefore, TDMs require more optimization on the architecture level, and more sample-efficient planning algorithms are needed to achieve faster real-time inference.
\begin{figure}[t!]
    \centering
    \includegraphics[width=0.51\textwidth, height=0.23\textwidth]{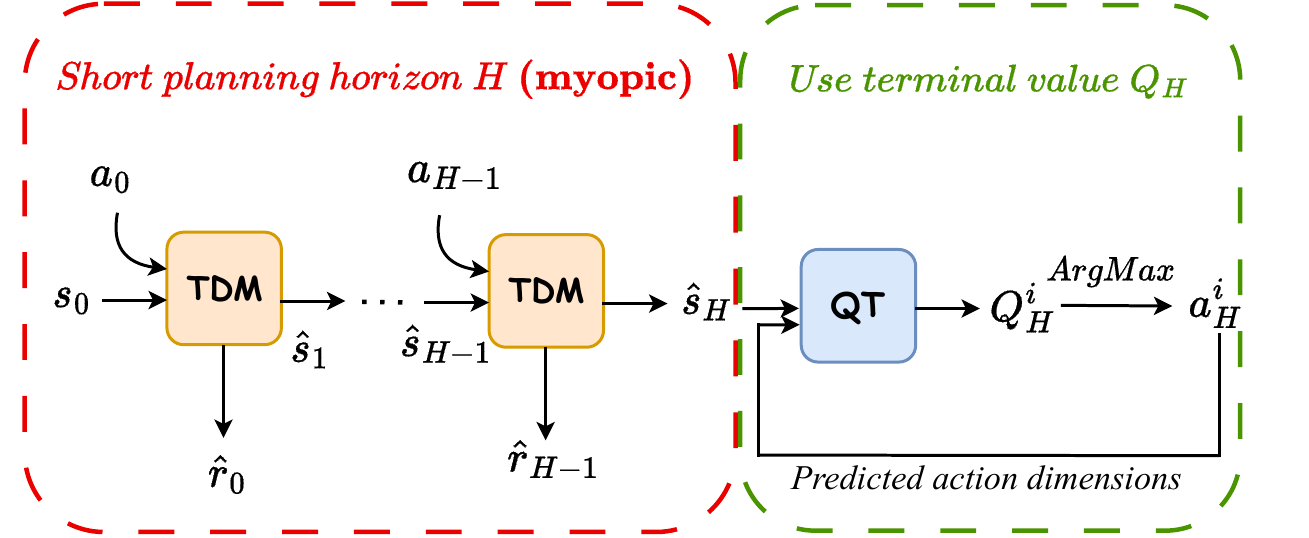}
    \caption{QT-TDM Inference: The learned TDM model plans for short planning horizon $H$, while the learned QT model estimates an autoregressive terminal value $Q^i_H$ for each action dimension $a^i_H$ which guides the planning beyond the myopic horizon.}
    \label{fig:0}
\end{figure}

To this end, we introduce QT-TDM, a model-based algorithm that combines the strengths of a TDM and a model-free Q-Transformer (QT) \cite{qtransformer}. Inspired by the TD-MPC algorithm \cite{tdmpc}, our proposed model achieves fast inference (as shown in Fig.~\ref{fig:0}) by combining a short planning horizon with a terminal value that is estimated by the Q-Transformer model which provides an estimate of a long-term return beyond the myopic planning horizon. Additionally, the sequence length is reduced by tokenizing the high-dimensional state space into a single token using a learned linear layer \cite{chen2021decision}, as opposed to the conventional per-dimension tokenization method \cite{generalistTDM}, \cite{janner2021offline}.

The advantages of QT-TDM are twofold. First, the modular architecture, consisting of two components (TDM and QT) that can be trained and used individually, facilitates the replacement and testing of its components. Second, the Transformer-based architecture, which incorporates GPT-like Transformers \cite{gpt}, allows for scalability through training with diverse offline datasets, thereby enhancing generalization.

In this paper, we evaluate the proposed QT-TDM for real-time continuous planning with Model Predictive Control (MPC) using diverse state-based continuous control tasks from two domains: DeepMind Control Suite \cite{dmc} and MetaWorld \cite{metaworld}. The results demonstrate the superior performance and sample efficiency of the QT-TDM model compared to baselines, while also achieving fast and computationally efficient inference. Our contributions can be summarised as follows:
\begin{itemize}
    \item We propose QT-TDM, a Transformer-based model-based algorithm consisting of two modules (QT and TDM) in a modular architecture.
    \item  QT-TDM addresses the slow and computationally inefficient inference associated with TDMs, while maintaining superior performance compared to baselines. 
\end{itemize}
\section{Related Work}

\subsubsection{Transformer Dynamics Model}
Motivated by the success of Transformers in sequence modeling tasks, there has been a lot of recent attention on using Transformers as dynamics models. One of the earliest attempts is \textit{TransDreamer} \cite{transdreamer} which as implied by the name is a modification of the Dreamer model \cite{dreamer}. TransDreamer replaces the Recurrent State-Space Model (RSSM) \cite{planet} with a Transformer State-Space Model (TSSM), improving TransDreamer's performance in long-term memory tasks. \textit{IRIS} \cite{iris} and \textit{TWM} \cite{twm} are two sample-efficient model-based agents that are trained inside the imagination of a Transformer-based world model. IRIS' world model consists of a discrete autoencoder \cite{discreteAE} as an observation model and a GPT-like Transformer \cite{gpt} as a dynamics model, while the world model of TWM consists of a variational autoencoder \cite{variationalAE} and a Transformer-XL \cite{transformer-xl}. Both models work with discrete action environments and they achieve impressive results on the Atari 100K benchmark.
\textit{Generalist TDM} \cite{generalistTDM} is the first attempt to use a learned TDM for continuous real-time planning with Model Predictive Control. Generalist TDM performs well in a single environment (i.e, specialist setting) and generalizes to unseen environments (i.e., generalist setting), in a few-shot and in zero-shot scenarios. Despite of its capabilities, it has two shortcomings. First, the training data is collected by an expert agent and not by its own interactions with the environment. Second, it suffers from slow inference because of the long-horizon planning and because of the design choices that are based on the Gato Transformer model \cite{gato} which uses the per-dimension tokenization scheme.

To overcome the above shortcomings, we introduce the QT-TDM model, which explores the environment to collect training data and has faster inference speed by shortening the planning horizon and utilizing the QT model \cite{qtransformer} to estimate a long-term return beyond the short-term planning horizon.

\subsubsection{Robotics Foundation Models}
Inspired by the success of Vision/Language Foundation Models \cite{vlm}, there remains significant potential for the development of specialized Robotics Foundation Models (RFMs). Foundation Models, primarily based on Transformer architectures, are pre-trained on large-scale datasets and exhibit remarkable zero-shot and few-shot generalization capabilities. Examples of RFMs are RT-2 \cite{rt2}, Q-Transformer \cite{qtransformer}, Gato \cite{gato} and PaLM-E \cite{palm}. All existing RFMs adopt a model-free (model-agnostic) approach. However, many researchers  argue that a model-based approach based on \textit{Foundation World Models (FWMs)} is a promising direction for addressing complex robotics challenges \cite{path}. While Generalist TDM \cite{generalistTDM} shows the potential of this direction, this work builds on it and further improves real-time planning capabilities and efficiency. In the future work section, we propose strategies to further advance QT-TDM toward the realization of FWMs.

\section{Background}

\begin{figure*}[t!]
    \centering
    \captionsetup[subfloat]{font=footnotesize,labelfont=sf,textfont=rm}
	\subfloat[TDM]{\includegraphics[width=0.55\textwidth]{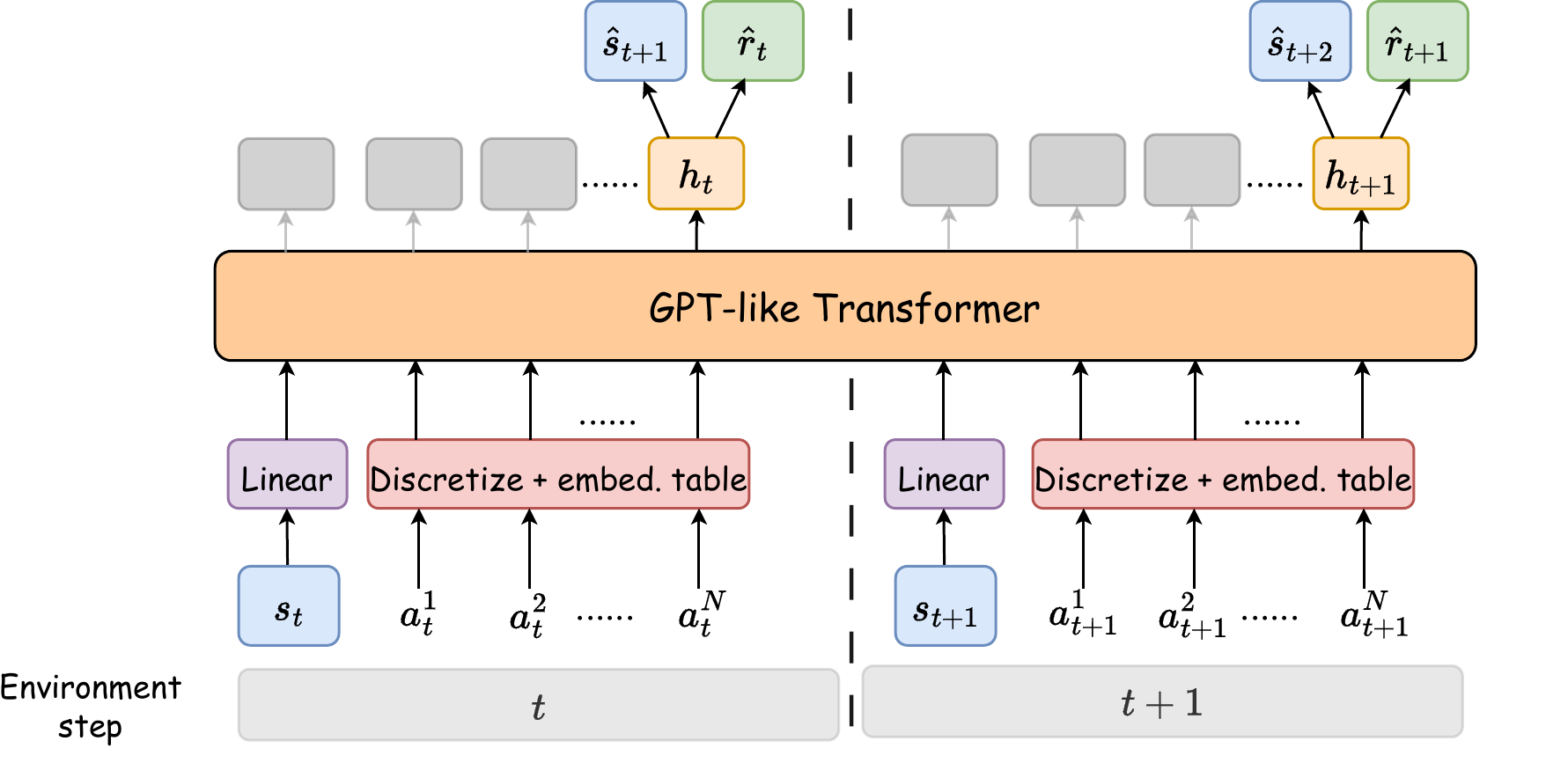}\label{fig:1a}}\quad\quad
	\subfloat[QT]{\includegraphics[width=0.30\textwidth]{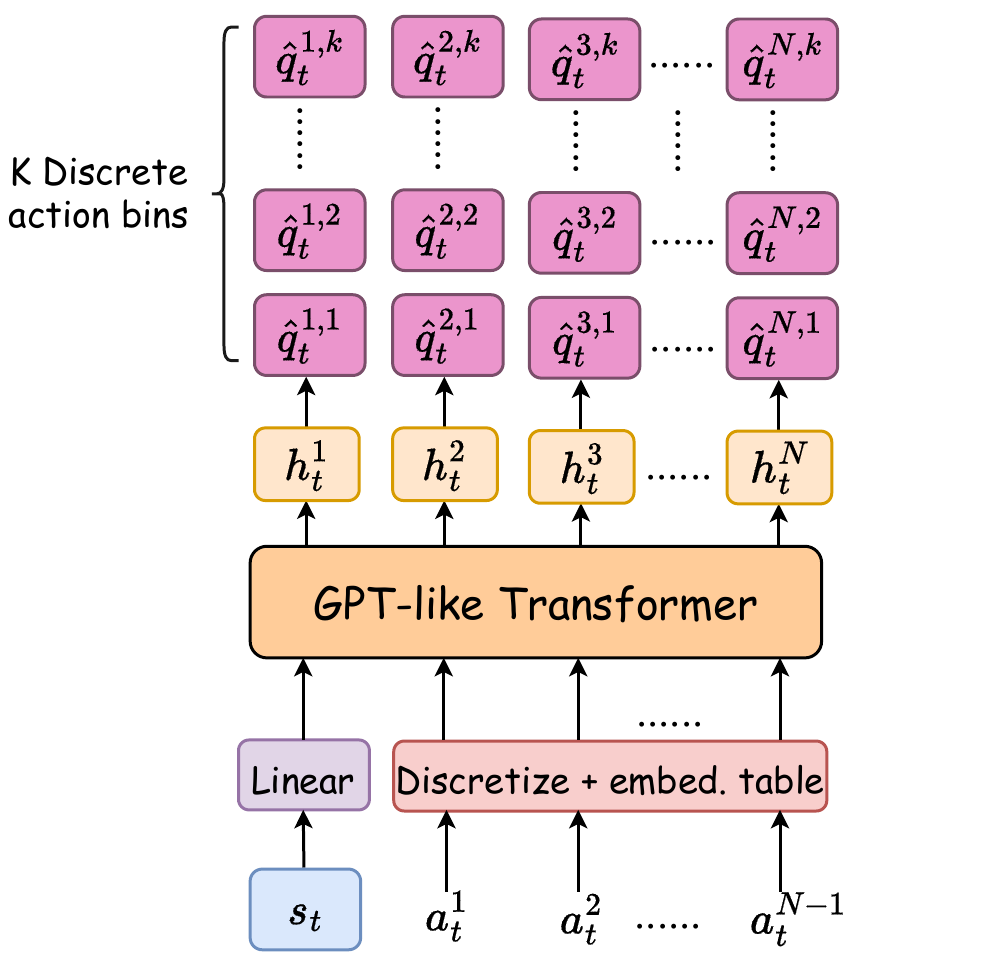}\label{fig:1b}} 
	
    \caption{QT-TDM Architecture, which consists of two modules: (a) TDM and (b) QT. Both modules have a GPT-like Transformer as a main component and share the same tokenization scheme. The state $s_t$ is tokenized into a single token using a learned linear layer. A per-dimension tokenization is performed for the $N$-dimensional action by discretizing each dimension independently into $K$ bins, then using an embedding table. The TDM module predicts the next state $\hat{s}_{t+1}$ and the reward $\hat{r}_t$ and is trained on $\mathit{L}$ sampled time steps (for brevity, we only show two time steps). The QT module predicts a Q-value for each action dimension $\hat{q}_t^{i,1:K}\hspace{0.5em}\forall i \in [1,...,N]$.}
    \label{fig:1}
\end{figure*}

\subsubsection{Reinforcement Learning}
We formulate the problem of continuous control as an infinite-horizon Markov Decision Process (MDP) that can be formalized by a tuple $(\mathcal{S}, \mathcal{A}, \mathcal{R}, \mathcal{T}, \gamma)$, where $\mathcal{S}$ is the state space, $\mathcal{A}$ is the continuous action space, $\mathcal{R}:\mathcal{S}\times\mathcal{A}\mapsto\mathbb{R}$ is a reward function, $\mathcal{T}:\mathcal{S}\times\mathcal{A}\mapsto\mathcal{S}$ is the transition function, and $\gamma \in [0,1]$ is a discount factor. The goal of reinforcement learning is to learn a policy $\Pi_{\theta}:\mathcal{S}\mapsto\mathcal{A}$ from interacting with the environment that maximizes the expected cumulative discounted reward $\mathbb{E}_{\Pi_{\theta}} [ \sum^{\infty}_{t=0} \gamma^{t}r_t]$. In this work, the policy $\Pi_{\theta}$ is derived from planning with a learned dynamics model.

\subsubsection{Model Predictive Control}
In control, learning $\Pi_{\theta}$ is formulated as a trajectory optimization problem, in which at each step $t$, optimal actions $a_{t:t+H}$ over a finite horizon $H$ are estimated to maximize the discounted sum of rewards:
\begin{equation}
    \Pi_{\theta}(s_t) = \arg \max_{a_{t:t+H}} \mathbb{E} \left[ \sum_{i=t}^{H} \gamma^i r_i \right],\label{eq:0}
\end{equation}
and the first action $a_t$ is executed. This method is known as \textit{Model Predictive Control (MPC)}. Eq.~\ref{eq:0} is not predicting long-term rewards beyond $H$. Consequently, incorporating  a value function of the terminal state $s_{t+H}$ provides an estimate of the long-term return, a method referred to as \textit{MPC with a terminal value} \cite{tdmpc}. An alternative approach, known as \textit{MPC with value summation} \cite{sample-efficiency3}, involves summing value functions over a finite horizon rather than summing rewards. In this work, we utilize Q-Transformer to estimate a terminal Q-value in a myopic planning horizon.

\subsubsection{Autoregressive Q-Learning}
Applying Q-learning with Transformers is challenging since Transformers require discretizing the action space into tokens to effectively apply the attention mechanism. Therefore, the standard Q-learning needs to be reformulated in order to be applied. In the Q-Transformer model \cite{qtransformer}, an autoregressive Q-learning formulation is proposed where each action dimension is treated as a separate time step. This way, each action dimension can be discretized individually, rather than discretizing the entire action space, thus avoiding the exponential growth in the discrete action space. An autoregressive discrete Q-function is employed which predicts a separate Q-value for each action dimension. 
Let $a_t = (a_t^1,...,a_t^N)$ be an $N$-dimensional action at time step $t$. The autoregressive Q-function predicts a Q-value for each action dimension $a_t^i$ that is conditioned on the state $s_t$ and the previous action dimensions $a_t^{1:i-1}$:
\begin{equation}
    Q(s_t,a_t^{1:i-1},a_t^i)\hspace{0.3em}\forall i \in [1,...,N].\label{eq:1}
\end{equation}  

To train the Q-function, a per-dimension Bellman update is defined as follows:
\begin{equation}
    Q(s_t,a_t^{1:i-1},a_t^i)\leftarrow
    \begin{cases}
        \max\limits_{a_t^{i+1}}Q(s_t,a_t^{1:i},a_t^{i+1})&\text{if}\hspace{0.2em}i < N\\
        r_t + \gamma \max\limits_{a_{t+1}^{1}}Q(s_{t+1},a_{t+1}^{1})&\text{if}\hspace{0.2em}i=N.\label{eq:2}
    \end{cases}
\end{equation}

The Q-targets of all action dimensions except the last one are computed by maximizing over the discretized bins of their subsequent dimension within the same time step. The Q-target of the last dimension is computed by the discounted maximization of the first dimension of the next time step plus the reward. The reward is only applied on the last dimension as it is observed after executing the whole action. In addition, the Q-values are only discounted between time steps (i.e., discount factor $\gamma$ is only applied for the last dimension), to ensure the same discounting as in the original MDP. 
The Q-Transformer model \cite{qtransformer} has been evaluated in an offline RL fashion with large-scale robotic sparse reward tasks. In this work, we utilize the Q-Transformer model in an online RL fashion to estimate a terminal Q-value in a short-horizon real-time planning task, in order to achieve faster planning.

\section{Methodology}
To resolve the trade-off between expressiveness and speed in TDMs,  we introduce QT-TDM, a model-based RL algorithm that captures the environment's dynamics by modeling trajectory data using a Transformer Dynamics Model and achieves fast inference speed by utilizing a terminal Q-value to guide a short-horizon planning (see Fig.~\ref{fig:0}). In this section, we first describe the architecture of our model, then the training procedure, and finally explain how to apply the Q-Transformer during planning.

\subsection{Architecture}
QT-TDM model shown in Fig.~\ref{fig:1} consists of two separated modules: Transformer Dynamics Model (TDM) and Q-Transformer Model (QT) \cite{qtransformer}. 

TDM is implemented as a GPT-like Transformer \cite{gpt} that computes a deterministic hidden state $h_t$ conditioned on the states and actions of past steps. We consider only the hidden state corresponding to the last action dimension, as it attends to all preceding action dimensions (see Fig.~\ref{fig:1a}; orange boxes vs. faded gray boxes).  Predictors for the next state and reward are conditioned on the hidden state which are implemented as multilayer perceptrons (MLPs). The model components are as follows:
\begin{subequations}\label{eq:3}
\begin{align}
    &\text{Hidden state:\hspace{1em}} h_t\hspace{0.75em} = \mathit{f}_{\theta}(s \leq t, a^{1:N} \leq t)\label{eq:3A}\\
    &\text{Transition:\hspace{2em}} \hat{s}_{t+1} = \mathit{g}_{\theta}(h_t)\label{eq:3b}\\
    &\text{Reward:\hspace{3em}} \hat{r}_{t}\hspace{1em} \sim \mathit{p}_{\theta}(\hat{r}_{t}|h_t),\label{eq:3c}
\end{align}
\end{subequations}
the reward model outputs the mean of a normal distribution.

The Q-Transformer model consists of a GPT-like Transformer and an autoregressive discrete Q-function that predicts a Q-value for each action dimension which is implemented as MLP. The Transformer computes a deterministic hidden state $h_t^i$ for each action dimension $a_t^i$ conditioned on the state $s_t$ and previous action dimensions $a_t^{1:i-1}$. 
The model components are as follows:
\begin{subequations}\label{eq:4}
\begin{align}
    &\text{Hidden state:\hspace{1em}} h_t^i\hspace{0.75em} = \mathit{f}_{\phi}(s_t, a_t^{1:i-1})\hspace{0.5em}\forall i \in [1,...,N]\label{eq:4A}\\
    &\text{Q-Value:\hspace{2.5em}} \hat{q}_t^{i,1:K}\hspace{1em} = \mathit{g}_{\phi}(h_t^i)\hspace{0.5em}\forall i \in [1,...,N],\label{eq:4B}
\end{align}
\end{subequations}
where $K$ is the number of discretized action bins.

Both models, TDM and QT, tokenize the input sequences in the same way. Let $s \in \mathcal{S}$ be an $M$-dimensional state and $a \in \mathcal{A}$ is an $N$-dimensional continuous action. We follow \cite{chen2021decision} in tokenizing the state $s$ into a single token obtained with a learned linear layer, rather than the conventional per-dimension tokenization \cite{generalistTDM}, \cite{janner2021offline} which increases the input sequence length. We perform a per-dimension tokenization for the $N$-dimensional continuous action $a = (a^1,a^2,....,a^N)$ by discretizing each dimension independently into $K$ uniformally-spaced bins, then invoking the token embedding from a learned embedding table. TDM takes as input a sequence of $\mathit{L}\times(N+1)$ tokens, where $\mathit{L}$ is time steps. QT takes as input a sequence of $N$ tokens as it ignores the last action dimension.
\RestyleAlgo{ruled}
\SetKwComment{Comment}{$\triangleright$\ }{}
\newcommand\mycommfont[1]{\footnotesize \textcolor{black}{#1}}
\SetCommentSty{mycommfont}
\SetKwInput{KwRequire}{Require}
\begin{algorithm}[t]
\caption{QT-TDM (Training)}
\label{alg:training}
\KwRequire{$\theta$: initialized TDM parameters\\
         $\phi$, $\bar{\phi}$: initialized QT parameters, EMA parameters
        \newline $\eta_d$, $\eta_q$: learning rates
        \newline $\mathcal{B}$, $\zeta$: replay buffer, EMA coefficient
        \newline $\mathit{L}$, $N$: sampled time steps, action dim.}
\For{$each\ training\ step$}{
            \nonl\textcolor{teal}{\textit{// Collect episode with QT-TDM and add to buffer}}
            
            $\mathcal{B}\gets\mathcal{B}\cup\{s_t, a_t,r_t, s_{t+1}\}_{t=0}^{T-1}$
            
            \For{$num\ updates\ per\ episode$}{
                    $\{s_t, a_t,r_t, s_{t+1}\}_{t=1}^{L}\sim\mathcal{B}$\Comment*[r]{Sample trajectory}
                    \nonl\textcolor{teal}{\textit{// Update Dynamics Model (TDM)}}
                    
                    \For{$t=1 ... L$}{
                                $h_t = \mathit{f}_{\theta}(s \leq t, a^{1:N} \leq t)$\Comment*[r]{Hidden state}
                                $\hat{s}_{t+1} = \mathit{g}_{\theta}(h_t)$\Comment*[r]{Transition}
                                $\hat{r}_t \sim \mathit{p}_{\theta}(\hat{r}_{t}|h_t)$\Comment*[r]{Reward}
                    }
                    
                    $\theta\gets\theta - \eta_d \grad_{\theta} \mathcal{L}_\theta^{Dyn}$\Comment*[r]{Equation~\ref{eq:8}}
                    \nonl\textcolor{teal}{\textit{// Update Q-Transformer (QT)}}

                    \For{$i=1 ... N$}{
                                $h_t^i = \mathit{f}_{\phi}(s_t, a_t^{1:i-1})$\Comment*[r]{Hidden state}
                                $\hat{q}_t^{i,1:K} = \mathit{g}_{\phi}(h_t^i)$\Comment*[r]{Q-Values}
                
                    }
                    $\phi\gets\phi - \eta_q \grad_{\phi} \mathcal{L}_\phi^{Q}$\Comment*[r]{Equation~\ref{eq:9}}
                    \nonl\textcolor{teal}{\textit{// Update Target Network}}
                    
                    $\bar{\phi} \gets (1 - \zeta)\bar{\phi} + \zeta\phi$;
            }

}

\end{algorithm}

\subsection{Training}
The dynamics model is trained in a self-supervised manner on segments of $\mathit{L}$ time steps sampled from the replay buffer $\mathcal{B}$. We minimize the sum of a mean-squared error transition loss and a negative log-likelihood reward loss:
\begin{equation}
    \mathcal{L}_{\theta}^{Dyn} = \sum^{\mathit{L}}_{t=1} \Bigl[\mathcal{\beta}_{1} \norm{\mathit{g}_{\theta}(h_t) - s_{t+1}}^2_2 - \mathcal{\beta}_{2} \ln{\mathit{p}_{\theta}(r_{t}|h_t)} \Bigr],\label{eq:8}
\end{equation}
where $\mathcal{\beta}_{1}$ and $\mathcal{\beta}_{2}$ are coefficients of the transition loss and the reward loss respectively.

The Q-Transformer model is trained by minimizing the Temporal Difference (TD) error loss defined by the per-dimension Bellman update \cite{qtransformer} in Eq. \ref{eq:2}
\begin{equation}
    \mathcal{L}_{\phi}^{Q} = Q_{\phi}(s_t,a_t) - Q_{\bar{\phi}}^*(s_t,a_t),\label{eq:9}
\end{equation}
where $Q_{\phi}(s_t,a_t)=\{\hat{q}^{i}_{t}\}_{i=1}^{N}$ consists of the predicted Q-values of all action dimensions, and $Q_{\bar{\phi}}^*$ are the target Q-values predicted by a Q-target network whose parameters are an exponential moving average (EMA) of the Q-network. We use smooth L1 loss \cite{smoothl1} as the TD-error which stabilizes training by avoiding exploding gradients. We follow \cite{qtransformer} in employing $n$-step return \cite{nstep} over action dimensions, and utilizing Monte Carlo return \cite{monte-carlo} only with sparse reward tasks (e.g., Reacher Easy), which helps accelerate learning. See Algorithm \ref{alg:training} for training pseudo code.

\RestyleAlgo{ruled}
\SetKwComment{Comment}{$\triangleright$\ }{}
\SetKwInput{KwRequire}{Require}
\begin{algorithm}[t]
\SetAlgoVlined
\caption{QT-TDM (Planning)}
\label{alg:planning}
\KwRequire{$\theta$, $\phi$: TDM parameters, QT parameters\\
        $\mu^0$, $\sigma^0$: initial parameters of $\mathcal{N}$
        \newline $J$, $J_{QT}$: num. of samples, num. of QT samples
        \newline $s_t$, $H$, $\mathcal{I}$: current state, len. of horizon, iterations}

\For{$n=1 ... \mathcal{I}$}{
        Sample $J$ action seq. from $\mathcal{N}(\mu^{n-1}, (\sigma^{n-1})^2 \mathbf{I})$
        
        Sample $J_{QT}$ action seq. using QT and TDM

        \nonl\textcolor{teal}{\textit{// Rollout trajectories and estimate total return $\mathcal{F}_{\varGamma}$}}

        \For{$all\ J+J_{QT}\ action\ sequences$}{
                \For{$t=0 ... H-1$}{
                        $h_t = \mathit{f}_{\theta}(s \leq t, a^{1:N} \leq t)$\Comment*[r]{Hidden state}
                        $\hat{s}_{t+1} = \mathit{g}_{\theta}(h_t)$\Comment*[r]{Transition}
                        $\mathcal{F}_{\varGamma} = \mathcal{F}_{\varGamma} + \gamma^{t}\mathit{p}_{\theta}(\hat{r}_{t}|h_t)$\Comment*[r]{Reward}
                }
            \nonl\textcolor{teal}{\textit{// Estimate the terminal Q-value using QT and add the value of last action dim. to $\mathcal{F}_{\varGamma}$}}

            $\mathcal{F}_{\varGamma} = \mathcal{F}_{\varGamma} + \gamma^{H}\max\limits_{a_{H}^{N}}Q_{\phi}(\hat{s}_{H},a_{H}^{N})$
        }
        Update $\mu^{n}$ and $\sigma^{n}$\Comment*[r]{Equation~\ref{eq:11}}
}
\textbf{return} $a_t \sim \mathcal{N}(\mu_t^{\mathcal{I}}, (\sigma_t^{\mathcal{I}})^2 \mathbf{I})$\Comment*[r]{First action is executed}

\end{algorithm}

\subsection{Planning}
We evaluate the proposed QT-TDM model on real-time planning with MPC, where inference speed needs to be taken into consideration. The inference time grows with the planning horizon $H$, the number of planning samples $J$, and the dimensionality of the environment $\mathcal{D}$. While Transformers serve as large, expressive, and robust dynamics models, they are not optimized for fast inference \cite{generalistTDM}. The per-dimension tokenization and the autoregressive token prediction lead to a slow inference over long horizons. To solve this issue and achieve faster inference, we use a short planning horizon and employ the Q-Transformer model to estimate a terminal Q-value \cite{tdmpc} that provides a long-term return beyond the short-term horizon.
During planning with MPC, we sample $J$ action sequences of length $H$ from a time-dependent multivariate diagonal Gaussian distribution initialized by $(\mu^0,\sigma^0)_{t:t+H}$. Then, trajectories are generated using rollouts from the learned dynamics model (TDM), and the total return $\mathcal{F}_{\varGamma}$ of a trajectory $\varGamma$ is computed as follows:
\begin{equation}\label{eq:10}
  \mathcal{F}_{\varGamma} =  \mathbb{E}_{\varGamma}\left[\gamma^{H}\max\limits_{a_{H}^{N}}Q_{\phi}(\hat{s}_{H},a_{H}^{N})+\sum^{H-1}_{t=0} \gamma^{t}\mathit{p}_{\theta}(\hat{r}_{t}|h_t)\right],
\end{equation}
where $Q_{\phi}(\hat{s}_{H},a_{H}^{N})$ is the terminal Q-value of the last action dimension $a_H^N$. The distribution $\mu^n$ and $\sigma^n$ at iteration $n$ are updated to the top-$k$ trajectories with the highest total returns $\mathcal{F}_{\varGamma}^\star$ as follows:
\begin{equation}\label{eq:11}
  \mu^n = \frac{\sum_{i=1}^{k} \Omega_i \varGamma_i^\star}{\sum_{i=1}^{k} \Omega_i}, \quad
\sigma^n = \sqrt{\frac{\sum_{i=1}^{k} \Omega_i \left( \varGamma_i^\star - \mu^n \right)^2}{\sum_{i=1}^{k} \Omega_i}},
\end{equation}
where $\Omega_i = e^{\tau \left( \mathcal{F}_{\varGamma,i}^\star \right)}$, $\tau$ is a temperature parameter controlling the sharpness of the weighting and $\varGamma_i^\star$ is the $i$th top-$k$ trajectory. After a fixed number of iterations $\mathcal{I}$, the planning procedure terminates and a trajectory is sampled from the final updated distribution. The first action is only executed as we plan at each decision step $t$. In addition to sampling from the Gaussian distribution, we also sample $J_{QT}$ action sequences from the learned Q-Transformer model. The planning procedure is summarized in Algorithm \ref{alg:planning} and shown in Fig.~\ref{fig:0}

\begin{figure}[t!]
\centering
    \captionsetup[subfloat]{font=footnotesize,labelfont=sf,textfont=rm}
    \subfloat[Walker Walk]{\includegraphics[width=0.140\textwidth]{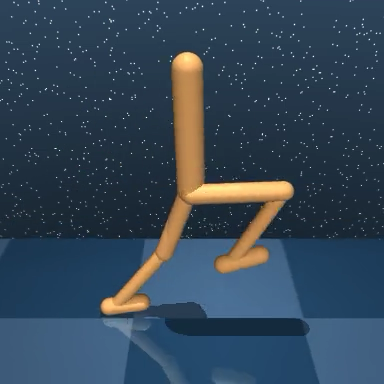}}\hspace*{0.1em}
    \subfloat[Cheetah Run]{\includegraphics[width=0.140\textwidth]{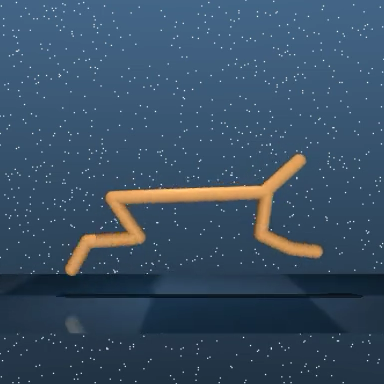}}\hspace*{0.1em}
    \subfloat[Reacher Easy]{\includegraphics[width=0.140\textwidth]{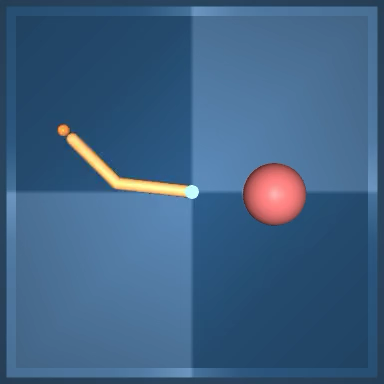}}
    
    \subfloat[Drawer Open]{\includegraphics[width=0.140\textwidth]{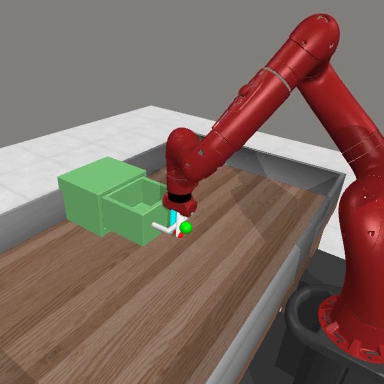}}\hspace*{0.1em}
    \subfloat[Plate Slide]{\includegraphics[width=0.140\textwidth]{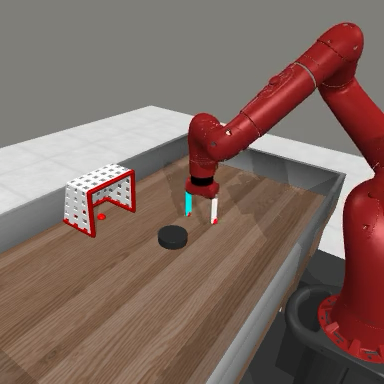}}\hspace*{0.1em}
    \subfloat[Reach Wall]{\includegraphics[width=0.140\textwidth]{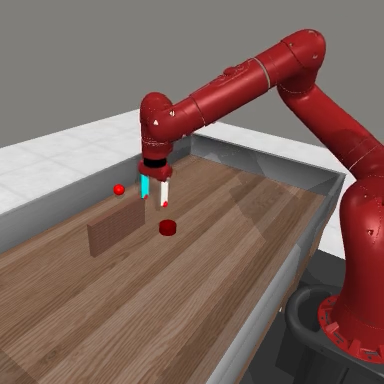}}
    
    \subfloat[Hammer]{\includegraphics[width=0.140\textwidth]{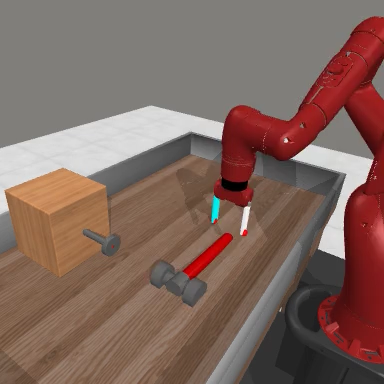}}\hspace*{0.1em}
    \subfloat[Door Unlock]{\includegraphics[width=0.140\textwidth]{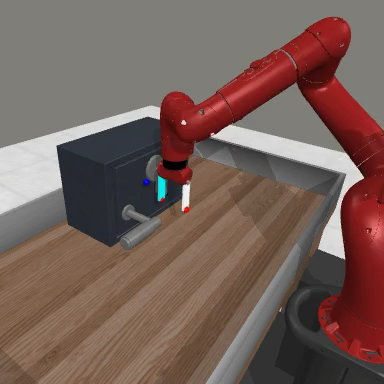}}\hspace*{0.1em}
    \subfloat[Button Press Wall]{\includegraphics[width=0.140\textwidth]{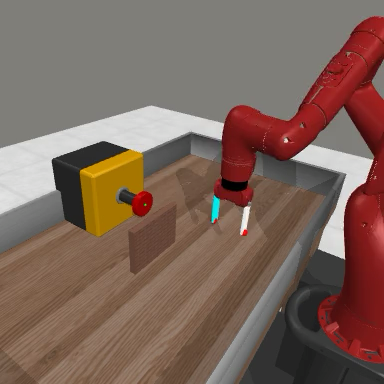}}
   
\caption{Continuous Control Tasks. Two locomotion tasks with high-dimensional action space (Walker and Cheetah) and one sparse reward task (Reacher) from DMC \cite{dmc}. Six robotic manipulation tasks (d)-(i) with various challenges from MetaWorld \cite{metaworld}.}
\label{fig:2}
\end{figure}

\section{Experiments}
\subsection{Description and Details}
\subsubsection{Benchmarks}
We evaluate the performance of QT-TDM model on diverse state-based continuous control tasks from two benchmarks: DeepMind Control Suite (DMC) \cite{dmc} and MetaWorld \cite{metaworld}. From DMC, we choose two high-dimensional locomotion tasks (\textit{Walker Walk} and \textit{Cheetah Run}) and a sparse reward task (\textit{Reacher Easy}). MetaWorld contains 50 different robotic manipulation tasks, and because of time and computational constraints, we choose six tasks with various challenges. All tasks are shown in Fig.~\ref{fig:2}.

\subsubsection{Baselines}
Since the \textit{Generalist TDM} \cite{generalistTDM} is the first Transformer-based model to perform continuous real-time planning, it serves as an eligible baseline. However, a comparison with it was not possible because its implementation is not publicly accessible. We compare the performance of QT-TDM against \textit{PlaNet} \cite{planet}, \textit{DreamerV3} \cite{dreamerV3} and its two individual modules \textit{(QT and TDM)} to serve as an ablation study as well. Both, PlaNet and DreamerV3 are model-based algorithms that use Recurrent State-Space Model (RSSM) as dynamics model. While PlaNet performs real-time planning with MPC, DreamerV3 performs background planning. Q-Transformer \cite{qtransformer} is a Transformer-based model-free algorithm that uses an autoregressive Q-Learning. We provide an extensive evaluation of QT on diverse tasks in an online RL scenarios.
TDM is a model-based algorithm that performs real-time planning but without the guidance from a terminal value function.

\subsubsection{Experimental Setup}
We list all the environment details for the tasks from the two benchmarks in Table~\ref{table:1} . For a fair comparison between the two Transformer-based model-based algorithms (QT-TDM and TDM), we use the same planning parameters shown in Table~\ref{table:2}. For the Recurrent-based model-based algorithm (PlaNet), we use its default planning parameters. All the compared models are evaluated after every $10$K environment steps averaging over $10$ episodes, except for DreamerV3, for which we use the final performance after convergence that we obtained from \cite{tdmpc2}.

\subsubsection{Computational Resources}
For each task, we trained our method and the baselines with 3 different random seeds. We ran our experiments with 6 Nvidia Quadro 6000 GPUs (24GB) using one GPU for one seed. For one DMC task, the total training and evaluation of our method takes on average 2 days while TDM takes 1.5 days. For one MetaWorld task, our method takes on average 4 days while TDM takes 3.5 days. The model-free QT takes 2 and 6 hours for training one DMC task and one MetaWorld task respectively.

\begin{table}[b!]
\caption{Environment details used across all methods for the two domains. We use action repeat of $4$ for DMC tasks except for Walker, where action repeat of $2$ is used.} 
\label{table:1}
\centering
\begin{tabular}{l c c}
\toprule
 & \textbf{DMC} & \textbf{MetaWorld}\\
\midrule
Episode length & $1000$ & $200$\\
Action repeat & $2$ / $4$  & $2$ \\          
Effective length & $500$ / $250$ & $100$\\
Environment steps & $500$K & $1$M\\
Performance metric & Reward & Success\\
\midrule
\multirow{3}{*}{Observation dim. $(M)$} & $6$ \textit{(Reacher)} & \\
                 & $17$ \textit{(Cheetah)} & $39$ \textit{(all tasks)}\\
                 & $24$ \textit{(Walker)} & \\
\midrule
\multirow{2}{*}{Action dim. $(N)$} & $6$ \textit{(Walker, Cheetah)} & $4$ \textit{(all tasks)}\\
            & $2$ \textit{(Reacher)} &  \\
\bottomrule
\end{tabular}
\end{table}

\begin{table}[b!]
\caption{MPC Planning parameters used for all tasks.} 
\label{table:2}
\centering
\begin{tabular}{l c c}
\toprule
 \textbf{Parameter}& \textbf{QT-TDM (ours) / TDM}& \textbf{PlaNet \cite{planet}} \\
\midrule
Initial parameters $(\mu^0,\sigma^0)$ & $(0,2)$ & $(0,1)$\\
Planning horizon $H$ & $3$ & $12$ \\
Num. of samples $J$ & $512$ & $1000$ \\
Num. of iterations $\mathcal{I}$ & $6$ & $10$ \\
Num. of top-$k$ trajectories & $64$ & $100$\\
\bottomrule
\end{tabular}
\end{table}

\begin{figure*}[t!]
\centering
\includegraphics[scale=0.64]{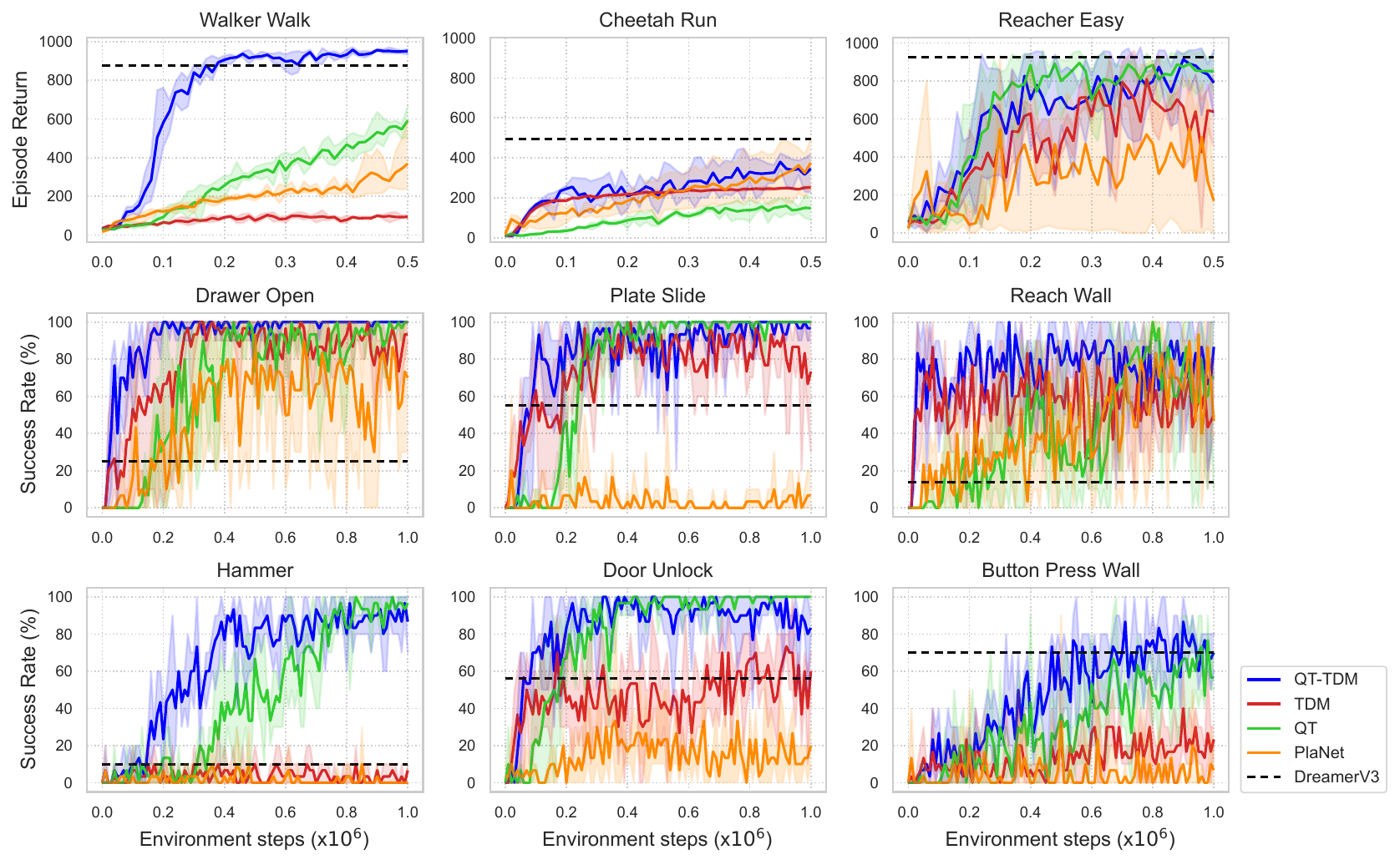}
\caption{Learning curves. Three tasks from DMC \textit{(top row)}, episode return as performance metric. Six tasks from MetaWorld \textit{(middle and bottom rows)}, success rate (\%) as performance metric. Mean over 3 seeds; shaded areas are standard deviations. For DreamerV3, we report the final performance from \cite{tdmpc2}. } 
\label{fig:3}
\end{figure*}

\subsection{Results}
Results for all 9 tasks from the two benchmarks are shown in Fig.~\ref{fig:3}. We summarize our findings as follows:

\subsubsection{Planning Efficiency}
\begin{figure}[t!]
    \centering
    \includegraphics[width=0.4\textwidth, height=0.3\textwidth]{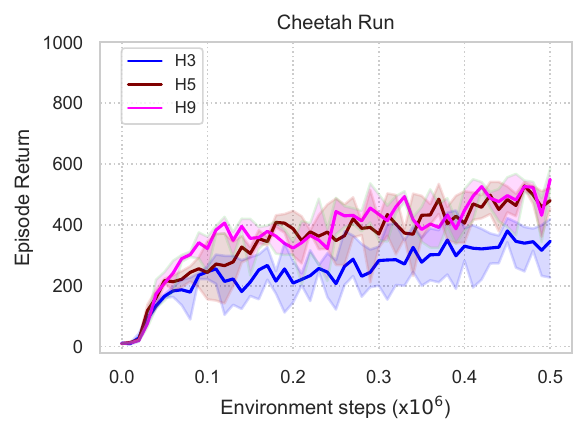}
    \caption{QT-TDM with different planning horizon $(H)$ on Cheetah Run task.}
    \label{fig:4}
\end{figure}

The two compared Transformer-based model-based algorithms (\textit{QT-TDM} and \textit{TDM}) perform real-time planning with a myopic planning horizon $(H = 3)$. However, QT-TDM relies on a learned terminal Q-value to guide the short-horizon planning. \textbf{\textit{In DMC tasks}}, TDM fails to solve the \textit{Walker} task, its learning stagnates at approximately $200$ returns after $100$K environment steps for the \textit{Cheetah} task, and it relatively solves the sparse reward \textit{Reacher} task at approximately $600$ returns. In contrast, our proposed QT-TDM model successfully solves all tasks, except for the \textit{Cheetah} task where it struggles a bit achieving approximately $400$ returns. We achieved improved results with planning horizons $H=5$ and $H=9$ as shown in Fig.~\ref{fig:4} but with the cost of higher inference time. \textbf{\textit{In MetaWorld tasks}}, while TDM struggles to solve hard tasks such as \textit{Hammer}, \textit{Door Unlock}, and \textit{Button Press Wall}, QT-TDM successfully solves all six tasks. QT-TDM outperforms TDM with only a $1.3\times$ increase in running time (e.g., from 1.5 days to 2 days for DMC tasks). This is more efficient than the over $2\times$ increase in running time required when extending the planning horizon $(H \ge 6)$. This demonstrates that our proposed QT-TDM achieves efficient real-time planning in terms of both performance and computational demands.

\subsubsection{Transformer vs. Recurrent}
Comparing our Transformer-based model against two Recurrent-based models highlights the superiority of TDMs in modeling dynamics. QT-TDM consistently outperforms PlaNet across all tasks, even though PlaNet utilizes a longer planning horizon (see table~\ref{table:2}). When compared to the state-of-the-art DreamerV3 which performs background planning, QT-TDM surpasses it in all MetaWorld tasks, while DreamerV3 achieves better performance in two DMC tasks (in Reacher Easy and Cheetah Run).

\subsubsection{Planning vs. Policy}
The compared model-free Q-Transformer selects actions with a value-based policy by maximizing Q-values over the discretized bins for all action dimensions. The QT model successfully solves all tasks from MetaWorld, but with less sample efficiency than our QT-TDM model. In DMC tasks with high-dimensional action spaces (\textit{Walker} and \textit{Cheetah}), QT was extremely sample-inefficient compared to QT-TDM. In the Walker task, QT achieves approximately 150 returns at 100K environment steps and 600 returns at 500K environment steps, compared with our proposed QT-TDM that achieves approximately 600 returns at 100K environment steps and 900 returns at 500K environment steps. It is expected that the model-based algorithm is more sample-efficient than its model-free counterpart \cite{sample-efficiency2}, \cite{sample-efficiency3}. Nevertheless, the results demonstrate that QT \cite{qtransformer} is a capable model-free algorithm that can perform effectively in both online and offline RL scenarios with sparse and dense rewards.

\subsection{Implementation}
Our GPT-like Transformer in both models (TDM and QT) is based on the implementation of \textit{minGPT} \cite{minGPT}. See Table~\ref{table:3} for the Transformer hyperparameters. The reward and next state predictors in TDM are implemented as 3-layer MLPs with dimension $512$, \textit{Leaky ReLU} activation, and $0.01$ dropout. We implement $2$ Q-functions in QT model as 2-layer MLPs with dimension $128$ and \textit{ReLU} activation. TD-targets are computed as the minimum of these $2$ Q-functions. Both models use Adam optimizer and Table~\ref{table:4} shows the optimization hyperparameters for TDM and QT. 

\begin{table}[b!]
\caption{Transformer hyperparameters.} 
\label{table:3}
\centering
\begin{tabular}{l c c}
\toprule
\textbf{Hyperparameter}& \textbf{TDM} & \textbf{QT}\\
\midrule
Input sequence & $\mathit{L}\times(N+1)$ tokens & $N$ tokens\\
Time steps $(\mathit{L})$ & $20$ & $1$\\
Discretize action bins $(K)$ & 256 & 256\\
Embedding dim. & $256$ & $128$\\
attention heads & $4$ & $8$\\
Num. of layers & $5$ & $2$\\
Embedding dropout & $0.1$ & $0.1$\\
Attention dropout & $0.1$ & $0.1$\\
Residual dropout & $0.1$ & $0.1$\\
\bottomrule
\end{tabular}
\end{table}

\begin{table}[b!]
\caption{Optimization hyperparameters.} 
\label{table:4}
\centering
\begin{tabular}{l c}
\toprule
\textbf{Hyperparameter} & \textbf{Value}\\
\midrule
\multicolumn{2}{c}{\textbf{TDM}}\\
\midrule
Batch size & $512$\\
Learning rate $(\eta_d)$ & \num{1e-4}\\
Weight decay & \num{1e-6}\\
Max gradient norm & $30$\\
Transition loss coef $(\beta_1)$. & $1.0$\\
Reward loss coef. $(\beta_2)$ & $2.0$\\
\midrule
\multicolumn{2}{c}{\textbf{QT}}\\
\midrule
Batch size & $512$\\
\multirow{2}{*}{Learning rate $(\eta_q)$} & \num{3e-4} (fixed) \textit{(DMC)}\\
                                          & \num{3e-4} (decay) \textit{(MetaWorld)}\\
Weight decay & \num{1e-6}\\
Max gradient norm & $20$\\
EMA coef. $(\zeta)$ & $0.005$\\
Target $(\bar{\phi})$ update freq. & $5$ \textit{(DMC)}, $10$ \textit{(MetaWorld)}\\
$n$-step return & $3$\\
Monte Carlo return & sparse reward tasks (Reacher)\\
Discount $(\gamma)$ & $0.98$ \\

\bottomrule
\end{tabular}
\end{table}
\subsection{Complexity Analysis}
We compare the complexity of our QT-TDM model against the Generalist TDM model \cite{generalistTDM} in terms of \textit{model size} (i.e., number of parameters) and \textit{inference speed}. Since the implementation of Generalist TDM is not available, we do not use quantitative measures for inference speed such as wall-time or FLOPs. Instead, we measure inference speed based on \textit{planning horizon $H$}, \textit{terminal value $Q^H$}, \textit{number of planning samples $J$} and \textit{number of tokens per timestep $\mathcal{T}$} (see Table~\ref{table:5}). Due to the per-dimension tokenization, Generalist TDM requires $(M+N+1)$ tokens per timestep: $M$ state tokens, $N$ action tokens, and one reward token. In contrast, QT-TDM requires only $(1+N)$ tokens per timestep by reducing the state tokens to a single token using a learned linear layer and by not using the reward token. Additionally, our model utilizes an $85$\% shorter planning horizon compared to the Generalist TDM model. However, our model leverages a terminal Q-value $N$ times, with one value for each action dimension. Despite the additional steps, the total planning steps required by our model \textit{($3\! +\! N$ steps)} remain fewer than those required by the Generalist TDM \textit{(at least $20$ steps)}. Consequently, QT-TDM achieves faster inference speed with $92$\% fewer parameters than Generalist TDM. The computational demands of handling a high number of samples $J$ can be mitigated by increasing parallelization (using multiple cores) \cite{generalistTDM}.

\begin{table}[h!]
\caption{Complexity related parameters.} 
\label{table:5}
\centering
\begin{tabular}{l c c}
\toprule
\multirow{2}{*}{\textbf{Parameter}} & \textbf{QT-TDM} & \textbf{Generalist TDM}\\ 
                   &  (ours)         & \cite{generalistTDM}\\
\midrule
Num. of parameters & $6$M & 77M \\
Planning horizon$(H)$ & $3$  & $20$ -- $100$ \\
Terminal value $(Q^H)$ & $N$ & Not used \\
Num. of samples $(J)$ & $512$  & $64$ -- $128$ \\
Num. of per timestep tokens $(\mathcal{T})$ & $1+N$ & $M+N+1$ \\
\bottomrule
\end{tabular}
\end{table}
\section{Conclusion and future work}

In this paper, we propose QT-TDM, a Transformer-based model-based algorithm that overcomes the slow and computationally inefficient inference associated with TDMs. 

\noindent
\textbf{Model size.} Although the QT-TDM model comprises two separate GPT-like Transformers, it has a relatively small number of parameters (6M) compared to other Transformer-based models such as Generalist TDM (77M). This helps mitigate the overfitting issue commonly encountered with high-capacity Transformers.

\noindent
\textbf{Inference speed.} The proposed QT-TDM achieves fast real-time inference by reducing the number of per timestep tokens and combining short-horizon planning with a learned terminal Q-value to guide the planning process. In addition to sampling random trajectories from a Gaussian distribution, we sample a small number of trajectories (only 24) from the learned Q-Transformer. Further improvements to inference speed could be achieved by reducing the number of random trajectories and incorporating more policy-sampled trajectories. We plan to investigate this strategy in future work. Another straightforward approach to increase inference speed is to train the Q-Transformer model using the imagined trajectories generated by TDM. The learned QT model can then be used to select actions during inference. This technique is referred to as \textit{learning inside imagination} \cite{dreamer}.  

\noindent
\textbf{Limitations}. First, QT-TDM relies heavily on the learned Q-values to guide the myopic planning horizon. However, learning a value function to approximate future returns is known to be unstable and prone to overestimation. We observe that the Q-Transformer model struggles to solve complex and hard-to-explore environments such as \textit{pick Place} and \textit{Shelf Place} from MetaWorld benchmark. As part of our future work, we plan to explore the use of an \textit{ensemble of Q-functions} instead of just two Q-functions \cite{tdmpc2} which helps mitigate the overestimation issue. Additionally, we intend to employ a \textit{categorical cross-entropy loss} as the TD-error rather than the traditional MSE regression loss, as it has been shown to be more effective and can accelerate the learning process \cite{stopregress}. Second, the use of per-dimension tokenization for the action space makes it difficult to scale to high-dimensional action spaces (e.g., humanoid robots) because it increases both the sequence length and the inference time.

\noindent
\textbf{Generalization.} In this work, we evaluate QT-TDM in online RL scenarios to solve single tasks \textit{(i.e., specialist agent)}. For future work, we plan to assess the generalization capabilities of the QT-TDM model \textit{(i.e., generalist agent)} by training it with large, diverse offline datasets and evaluating its performance in unseen environments through few-shot and zero-shot scenarios.

\noindent
\textbf{Pixel observations.} In this work, we experiment exclusively with state-based environments. We plan to extend our approach to pixel-based environments in future work by developing an observation model such as ViT \cite{vit} or discrete autoencoder \cite{discreteAE}.

\section*{Acknowledgment}
The authors would like to thank Ozan Özdemir for his suggestions during the revision phase.


\IEEEtriggeratref{8}

\bibliographystyle{IEEEtran}
\bibliography{IEEEabrv,references.bib}

\end{document}